%% file: sample-sigconf.tex
\documentclass[sigconf]{acmart}

\usepackage{booktabs} 
\usepackage{graphicx}
\usepackage{algorithm}
\usepackage{mdframed}
\usepackage{fancybox}
\usepackage{nicefrac}

\usepackage{multirow}

\usepackage{xcolor}
\usepackage{fancybox}

\usepackage[most]{tcolorbox}

\definecolor{ShadowColor}{RGB}{120,120,120}

\makeatletter
\newcommand\Cshadowbox{\VerbBox\@Cshadowbox}
\def\@Cshadowbox#1{%
	\setbox\@fancybox\hbox{\fbox{#1}}%
	\leavevmode\vbox{%
		\offinterlineskip
		\dimen@=\shadowsize
		\advance\dimen@ .2\fboxrule
		\hbox{\copy\@fancybox\kern.2\fboxrule\lower\shadowsize\hbox{%
				\color{ShadowColor}\vrule \@height\ht\@fancybox \@depth\dp\@fancybox \@width\dimen@}}%
		\vskip\dimexpr-\dimen@+0.2\fboxrule\relax
		\moveright\shadowsize\vbox{%
			\color{ShadowColor}\hrule \@width\wd\@fancybox \@height\dimen@}}}
\makeatother

\setcopyright{rightsretained}

\acmConference[GECCO '17]{the Genetic and Evolutionary Computation Conference 2017}{July 15--19, 2017}{Berlin, Germany}
\acmYear{2017}
\copyrightyear{2017}

\begin{document}
	

\title{A Probabilistic Linear Genetic Programming with Stochastic Context-Free
Grammar for solving Symbolic Regression problems}

\author{L\'{e}o Fran\c{c}oso Dal Piccol Sotto}
\affiliation{%
  \institution{Federal University of S\~{a}o Paulo (UNIFESP)}
  \streetaddress{Avenida Cesare Mansueto Giulio Lattes, 1201}
  \city{S\~{a}o Jos\'{e} dos Campos} 
  \state{SP, Brazil} 
  \postcode{12231-280}
}
\email{leo.sotto@unifesp.br}

\author{Vin\'{i}cius Veloso de Melo}
\affiliation{%
  \institution{Federal University of S\~{a}o Paulo (UNIFESP)}
	\streetaddress{Avenida Cesare Mansueto Giulio Lattes, 1201}
	\city{S\~{a}o Jos\'{e} dos Campos} 
	\state{SP, Brazil} 
	\postcode{12231-280}
}
\email{vinicius.melo@unifesp.br}


\begin{abstract}
Traditional Linear Genetic Programming (LGP) algorithms are based only on the selection mechanism to guide the search. Genetic operators combine or mutate random portions of the individuals, without knowing if the result will lead to a fitter individual. Probabilistic Model Building Genetic Programming (PMB-GP) methods were proposed to overcome this issue through a probability model that captures the structure of the fit individuals and use it to sample new individuals. This work proposes the use of LGP with a Stochastic Context-Free Grammar (SCFG), that has a probability distribution that is updated according to selected individuals. We proposed a method for adapting the grammar into the linear representation of  LGP. Tests performed with the proposed probabilistic method, and with two hybrid approaches, on several symbolic regression benchmark problems show that the results are statistically better than the obtained by the traditional LGP.
\end{abstract}

%
%


\maketitle

\input{samplebody-conf-v3}

\bibliographystyle{ACM-Reference-Format}


\end{document}

%% file: samplebody-conf-v3.tex
\section{Introduction}
\label{sec:Introduction}

Evolutionary Algorithms (EAs,~\cite{Back:1996:EAT:229867}) are stochastic methods that use principles of natural evolution to search for solutions to optimization problems. Their search is based on random modifications on the individuals of a population, performed by mutation and crossover operators. In traditional EAs, the only mechanism that guides the search to promising regions is the selection, which is based on the fitness of the individuals. Other than that, EAs have no knowledge on the search space.

This non-informed search issue motivated the design of Estimation of Distribution Algorithms (EDAs)~\cite{Hauschild2011111}. EDAs are derived from Genetic Algorithms (GAs)~\cite{Sastry2005}, but use a probability model to sample individuals. At each generation, individuals are selected from the population and used to update the model. This way, the probability of sampling good solutions is increased and the search concentrates on promising regions of the search space. The aim is not only to increase the search efficiency but also its efficacy by solving previously unsolvable problems. Therefore, probability models can provide mechanisms to largely improve the performance of search algorithms, being a very relevant research topic.

A very popular example of EA used to design computer programs is the Genetic Programming algorithm (GP)~\cite{Koza:1992:GPP:138936}. In this research field, EDAs are usually called Probabilistic Model Building Genetic Programming algorithms (PMB-GP) \cite{Kim2014}, and there are several successful works showing that PMB-GP outperforms traditional GP~\cite{Salustowicz:1997:PIP:1326744.1326747, Yanai04programevolution}.

In this work, we propose a PMB-GP to improve a GP variant. We developed a probability model for the Linear Genetic Programming (LGP) algorithm~\cite{onLGPBrameier} and evaluated the resulting technique on Symbolic Regression (SR) problems. As far as we know, there is no grammar-based LGP, making this work novel. The model we chose for this work is the Stochastic Context Free Grammar (SCFG). LGP was chosen over GP because it presents interesting characteristics that can make it perform better than GP; such characteristics will be explained later in this paper.

The contributions of this work are: 1) introducing SCFG into LGP; 2) the development of a method for updating an SCFG and sampling LGP individuals from it; 3) testing the proposed algorithm on SR problems; 4) the development of a hybrid method to retain LGP features such as non-effective code, code reuse, and mutations; 5) a brief analysis of the impact of retaining the LGP features on the SR results.

The rest of the work is organized as follows. Sections \ref{subsec:PMB-GP} and~\ref{subsec:Linear-Genetic-Programming} present works related to PMB-GP and the background needed for understanding the proposed technique. The proposed technique is explained in detail in Section~\ref{subsec:Proposed-Technique}. Section~\ref{sec:Experiments} presents the experimental setup and the experimental results using a simple grammar and a more complex one, along with discussions. Finally, the conclusion and future works are discussed in Section~\ref{sec:Conclusion-and-Future}.

\section{Probabilistic Model Building Genetic Programming}
\label{subsec:PMB-GP}

Over the last years, many techniques and models have been proposed to incorporate domain knowledge into the tree-based GP search. Below we briefly describe some of them.

Probabilistic Incremental Program Evolution (PIPE)~\cite{Salustowicz:1997:PIP:1326744.1326747} uses Probability Prototype Trees~(PPTs). A PPT is a standard tree containing the maximum size an individual can reach. Each node has a table with the probability of that specific node assuming each terminal or non-terminal allowed for the given problem. However, nodes are independent from each other.

In \cite{1299866}, PPT is extended to model the conditional probability of each node with its parent by means of a Bayesian Network. The algorithm was called Estimation of Distribution Programming~(EDP). In \cite{Yanai04programevolution}, EDP is extended to a hybrid version using crossover and mutation, improving the performance.

A greedy search combined with the MDL (Minimum Description Length) metric is used in \cite{Sastry2003} to group nodes of a PPT. This grouping strategy makes the model multivariate through the calculation of the joint probability distribution. It can automatically identify non-overlapping building blocks (BBs) to improve recombination and reduce the chance of breaking good building blocks. 

As previously introduced, grammars have also been studied as a model for PMB-GP.  Shan et al. proposed Program Evolution with Explicit Learning~(PEEL)~\citep{PEEL}, which uses a stochastic grammar in which the Left-Hand Sides (LHS) also consist of the depth and the relative location of the given tree node. The rules are refined along the evolution process and updated by Ant Colony Optimization~(ACO), in which a pheromone value is maintained for each derivation to inform preferable paths. A similar approach is used in~\cite{tanev:2004:geccowks}.

A different approach for learning grammars is used in \cite{1330895}. At each generation, a very specific SCFG is learned for each best-fitted individual. The rules are then merged in order to become more general until the grammar can no longer be improved. The merging is done by a greedy search using the MML (Minimum Message Length) metric. The work of Bosman and De Jong \cite{Bosman04grammartransformations} uses the same search strategy for adding new rules (subfunctions) to the grammar employed in their algorithm.

Wong et al. \cite{6900423} proposed the GBBGP (Grammar-Based Bayesian Genetic Programming) that uses a Bayesian Network associated with each rule of an SCSG. The network models the probability of choosing a derivation based on the parent node, sibling nodes, and other context elements. An extension of that technique is proposed in \cite{Wong:2014:GGP:2576768.2598256}, where a Bayesian Network Classifier is used to derive a probability distribution for each rule.

Regarding the EA, perhaps the most similar work to the present one is~\cite{Poli:2008:LEG:1792694.1792713}, where N-grams are used in an LGP system. In an \textit{N-gram}, the probability of a random variable at position $i$ is conditioned to the values of the $N-1$ last positions. The authors report that their system was more scalable than LGP, being able to solve more difficult problems and more frequently.  


In the next section we introduce the EA used in this work.

\section{Linear Genetic Programming}
\label{subsec:Linear-Genetic-Programming}

Linear Genetic Programming (LGP) is a variant of GP that represents individuals linearly as sequences of instructions~\cite{onLGPBrameier}. The result of each instruction is stored into a register. The instruction consists of an operator that acts on operands, which can be input data, constants, or registers. An example of such a program for the SR task is shown in Figure~\ref{fig:Example-LGP-program}.

\begin{figure}[h]
\begin{minipage}[t]{0.45\columnwidth}

\begin{tcolorbox}[enhanced jigsaw, drop fuzzy shadow=ShadowColor, colback=gray!5]
~~~~~{0:~r{[}1{]}~=~x~{*}~1}{\par}
~~~~~{1:~r{[}2{]}~=~x~{*}~r{[}1{]}}{\par}
~~~~~{2:~r{[}0{]}~=~r{[}2{]}~+~3}{\par}
~~~~~~{3:~r{[}4{]}~=~r{[}2{]}~+~r{[}1{]}}{\par}
\end{tcolorbox}	
\end{minipage}
\caption{Example of an LGP program, representing the formula $f(x)=x^{2}+x$. The first number of each line is the instruction identifier for further reference. The value stored in the last register (instruction 3) is the program's output.}
\label{fig:Example-LGP-program}
\end{figure}

The linear representation introduces two features: non-effective code and code reuse. The non-effective code are instructions attributed to registers that are not used to compute the final result. Instruction~2 in Figure~\ref{fig:Example-LGP-program}, for instance, is non-effective because it is not used afterwards. These instructions help to increase the number of neutral variations (variations that do not change the result of the program) and can make individuals more flexible - a genetic operation on a non-effective instruction can make it effective.

The other feature - code reuse - can be exemplified by instruction~0 in Figure~\ref{fig:Example-LGP-program}. It is used by instruction~1 and again by instruction~3. This feature is useful if the same result must be used more than once in the same program, helping to evolve simpler individuals.

Crossover in traditional LGP is as in GA (block swap between parents), while mutation can be of two types: macro and micro-mutations. Macro-mutations change a complete instruction either deleting, inserting, or substituting a random instruction. On the other hand, micro-mutations change one element of an instruction like the destination register, the operator, or an argument. Given the existence of non-effective code and neutral variations, operators can be made effective, that is, when possible, change only the effective code of the individual.

The LGP implementation used in this work, called \textit{effmut}, uses only effective mutations, as reports show that this configuration performs the best~\cite{onLGPBrameier,EffectiveLGP}. It is a steady-state algorithm, as explained in detail in \cite{onLGPBrameier}. At each generation, two tournaments are carried out, yielding two winners and two losers. A copy of the winner replaces the loser in the population, while the original winner undergoes mutation according to the mutation rates. More details about the operators are found in~\cite{onLGPBrameier}.

\section{The Proposed Technique}
\label{subsec:Proposed-Technique}


In this section, we describe the model used in our approach, the algorithm to sample individuals, the rule used to update the model probabilities, and the hybrid algorithms.

\subsection{The Model}
\label{subsec:The-Model}

An SCFG is a CFG (Context Free Grammar) in which each rule has a probability distribution associated with it. Such a grammar  G can be defined by a quintuple $G=(T,NT,S,R,P)$ where: $T$ is the set of terminal symbols; $NT$ is the set of non-terminal symbols; $S$ is the start symbol; $R$ is the set of production rules; and $P$ is the set of probabilities on production rules.

\begin{figure}[htb]

\begin{tcolorbox}[enhanced jigsaw, drop fuzzy shadow=ShadowColor, colback=gray!5]
	{\footnotesize{}Exp~:=~Exp~+~Term $\mid$ Exp~-~Term $\mid$ Term $\mid$ }\textit{\footnotesize{}probs}{\footnotesize{}~0.5~0.25~0.25}{\footnotesize \par}
	{\footnotesize{}Term~:=~Term~{*}~Factor~$\mid$~Term~/~Factor~$\mid$~Factor~$\mid$~}\textit{\footnotesize{}probs}{\footnotesize{}~0.5~0.0~0.5}{\footnotesize \par}
	{\footnotesize{}Factor~:=~(Exp)~$\mid$~Num~$\mid$~X~$\mid$~}\textit{\footnotesize{}probs}{\footnotesize{}~0.1~0.1~0.8}{\footnotesize \par}
	{\footnotesize{}Num~=:~1~$\mid$~2~$\mid$~3~$\mid$~4~$\mid$~5~$\mid$~6~$\mid$~7~$\mid$~8~$\mid$~9~$\mid$~}\textit{\footnotesize{}probs}{\footnotesize{}~0.11~0.11~0.33~0.0~0.0~0.11}
	{\footnotesize{}~0.22~0.11~0.0}{\footnotesize \par}
	{\footnotesize{}X~:=~$x_1$~$\mid$~$x_2$~$\mid$~}\textit{\footnotesize{}probs}{\footnotesize{}~0.8~0.2}{\footnotesize \par}
\end{tcolorbox}

\caption{Example SCFG for solving symbolic regression problems. Each rule has a probability distribution associated with it after the keyword \textit{probs}.}
\label{fig:Example-SCFG}
\end{figure}

In this work, we are assuming that the probabilities for each rule are independent. Hence, given a production $A:=B|C|D$, the probabilities of choosing to derive \textit{B}, \textit{C}, or \textit{D} depend neither upon the parent derivation nor the depth in the current derivation tree. Figure~\ref{fig:Example-SCFG} shows an example of an SCFG with the rules and the corresponding probabilities for each derivation.

\subsection{Sampling Individuals}
\label{subsec:Sampling-of-Individuals}

To sample an individual from the SCFG, we use a recursive leftmost derivation to construct a  syntax tree; the difference being that the final result of the process is a sequence of instructions instead of a tree. The resulting program is a linear representation of what would be a syntax tree, and has only \textit{effective code} and \textit{no reuse}. While the tree is a functional paradigm where the results are passed from the bottom to the top, the linear representation is the procedural paradigm. Thus, it may be seen as an upside-down tree because the arguments for a function must be instantiated previously in the code.

\begin{figure*}[htbp]
	\begin{tcolorbox}[enhanced jigsaw, drop fuzzy shadow=ShadowColor,   interior hidden]
	\center
	\includegraphics[scale=0.5]{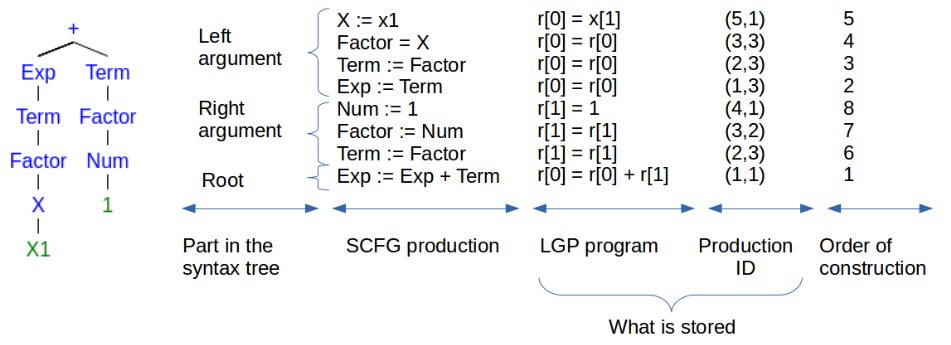}
	\end{tcolorbox}
	\caption{In the left, one has the syntax tree of $f(x) = x + 1$ using the grammar shown in Figure~\ref{fig:Example-SCFG}. In the right, one has the LGP program that represents the tree, along with the parts of the tree that corresponds to each part of the program, and the associated grammar production. The order of construction is the order in which the recursive algorithm builds the program. The production ID $(i,j)$ identifies which production is associated with each instruction, where $i$ is the number of the rule and $j$ is the number of the associated production.}
	\label{fig:exemplo-amostragem}
\end{figure*}

One begins by sampling from the start symbol of the grammar, which is represented by the last instruction of the program. If the sampled production requires the further derivation of non-terminals, we  call the function recursively (leftmost). The sequence of instructions for the leftside argument of a binary operator comes first in the individual; thus, the same store register cannot be used later to store the results of the rightside argument. Figure \ref{fig:exemplo-amostragem} shows an example of a derivation tree and the equivalent LGP program.

In the example, it can be noticed that there are repeated instructions. While the derivation is a non-terminal, a register receives itself because one must keep track of where an instruction came from. This way, one can update the correct production rule. This will be explained further.

As a consequence of the chosen representation, the maximum depth of a full binary tree equivalent to the sampled program is limited by the maximum number of registers. Given that each inner node of the equivalent tree corresponds to an instruction in the program, one has:

\begin{equation}
I=\sum_{i=0}^{D-1}2^{i}=2^{D}-1,
\end{equation}

\noindent where \textit{I} is the number of instructions and \textit{D} is the tree depth (starting with 0). Thus, a full binary tree with depth 9 would need 9 registers and 511 instructions to be represented by an LGP program using our sampling algorithm. Nevertheless, individuals sampled from the grammar tend to be much shorter than that.

\subsection{Update Rule}
\label{subsec:Update-Rule}

At each generation, the probability distribution associated with each rule of the grammar is updated towards the best individuals. That way, individuals sampled from the updated grammar are supposed to be better fitted to the given problem than if they are generated randomly. The update rule used in the proposed algorithm is similar to the one used in  PBIL~\cite{Baluja:1994:PIL:865123}.

First, the \textit{N} best individuals are selected from the population. As seen in Figure \ref{fig:exemplo-amostragem}, each instruction has an ID indicating \textit{which production of the grammar was used to derive it}. With such information from the selected individuals, one calculates a proportion of use of productions for each distribution. Then, the probability distribution of each rule is updated according to the following formula:

\begin{equation}
Prob_{ij}^{g+1}=(1-\alpha)*Prob_{ij}^{g}+\alpha*Prop_{ij}^{g},
\end{equation}

\noindent where $\alpha$ is the learning rate that ranges from $0$ to $1$, $g$ is the current generation, $i$ is the rule index, $j$ is the production index, $Prob$ is the probability distribution, and $Prop$ is the proportion of use. The proportion calculation is carried out for all instructions considering all $N$ selected individuals at the same time.

As it stands, the proposed Grammar-Based LGP (GB-LGP) method of updating the model and resampling the population at each generation without using the genetic operators is not different from a GP with the same kind of grammar (EDA style). The only difference here is that the tree is represented linearly - no LGP extra feature is present. In order to introduce non-effective code, code reuse, and make use of the macro and micro-mutation operators, a hybrid scheme was developed.

\subsection{The Hybrid Approach}
\label{subsec:The-Hybrid-Approach}

We designed a hybrid approach that combines the mutation operators of LGP with the resampling from SCFG. The role of the hybrid approach is to introduce, via mutation operators, non-effective code and code reuse, and to add the benefits of the LGP evolution into the technique. Also, the running time of the algorithm is reduced when the resampling rate is lowered, as it is a costly procedure.

Two hybrid schemes were investigated in this work: 1) Resampling the entire population at each $s$ generations and applying mutations in the other ones (Hybrid GB-LGP v1); 2) At each generation, resample half of the population and generate the other half by mutations (Hybrid GB-LGP v2). In this last scheme, a tournament is performed, the winner is mutated, and the original winner along with the resulting individual of the mutation are passed to the next generation.

An issue arises because one must keep track of the productions used to generate the instructions.  When an individual's instruction is mutated, part of the grammar productions associated to it may no longer be valid, misleading the update rule of the distributions. This may change not only the instruction being mutated, but also the production of its parent or children. To simplify the present implementation, we ignore this cascade effect and focus only on the instruction being mutated. However, it is paramount to deal with such effect to avoid breaking the model.

Macro and micro-mutations can change the production of an instruction to one that whether involves the identity operator (such as $r[0]=r[0]$) or not. When mutation does not involve the identity operator, one simply replaces the production that is currently associated to the instruction with the production that has the selected operator.

On the other hand, when a production is changed to another one that involves the identity operator, more than one option is possible for the new production. For instance, in the grammar in Figure~\ref{fig:Example-SCFG}, the possibilities to generate a terminal node (argument) are in the following path: \textit{Exp := Term, Term := Factor, Factor := (Exp), Factor := Num, Factor := X}, and the productions for rules $Num$ and $X$. For this case, one checks the content of the argument register of the identity operator. For instance, in the sequence $r[1]=1;~r[1]=r[1];~r[1]=r[1]$, the argument register is $r[1]$ and its content is $1$. Based on the production used to generate this instruction, one knows which production to associate to the resulting mutated instruction.

\section{Experiments}
\label{sec:Experiments}

In this section, we present the experimental analysis of the proposed algorithms on well-known SR benchmark functions of distinct difficulty levels. The experiments were elaborated to investigate the performance of the proposed techniques against the \textit{effmut} baseline, with the objective of outperforming it. Thus, in this study, we are not interested in comparing our methods with others than \textit{effmut}.

\subsection{Experimental Setup}
\label{subsec:Experimental-Setup}

The configuration of algorithms \textit{effmut}, GB-LGP, Hybrid1, and Hybrid2 are presented in Table~\ref{tab:Configuration-tested-techniques}. These values were either suggested by previous works on LGP \cite{onLGPBrameier,EffectiveLGP} or chosen empirically without fine-tuning. The operators allowed for \textit{effmut} are the same that appear in the grammar for each of the two experiments. However, \textit{effmut} does not use a grammar. GB-LGP and the hybrids need more registers to allow for larger trees to be derived, as explained before.

\begin{table}[h]
\center%
\begin{tabular}{c|c}
\hline 
\textbf{Parameter} & \textbf{Value}\tabularnewline
\hline 
\hline 
Initial Program Size & 20{*}\tabularnewline
Maximum Program Size & 200{*}\tabularnewline
Allowed Registers & 8{*}, 13{*}{*}\tabularnewline
Population Size & 100\tabularnewline
Generations & 100\tabularnewline
Elite & 1\tabularnewline
Tournament Size & 2\tabularnewline
Macro-Mutation Rate & 0.75\tabularnewline
Insertion & 0.66\tabularnewline
Deletion & 0.33\tabularnewline
Micro-Mutation Rate & 0.25\tabularnewline
$N$ & 3{*}{*}\tabularnewline
$s$ & 2{*}{*}\tabularnewline
$\alpha$ & 0.1{*}{*}\tabularnewline
\hline 
\end{tabular}
\caption{Configuration of the tested techniques. The symbol {*} means that the configuration is valid only for \textit{effmut}, while {*}{*} is for GB-LGP and the hybrids.}
\label{tab:Configuration-tested-techniques}
\end{table}

First, we tested the algorithms on simpler benchmark functions that require only sum and multiplication to be solved, using an equally simple grammar. We later tested them using a grammar with more options on more complex functions.

We performed 100 independent runs and compared the results using 
the Median of the Mean Absolute Error~(MMAE), the Median Absolute Deviation~(MAD)\footnote{$MAD = median |V_{i}- median (V)|, i=1,...,L$, where $V$ is an array of numerical values and $L$ is its length.}, and the success rate. A solution is successful if its mean error is less than $1e\textrm{-}05$. For the statistical comparison we employed the  Pairwise Wilcoxon's Rank-Sum Test at significance level $\alpha=0.05$. 

We implemented all algorithms in Python and ran 
the experiments on a system environment with an Intel(R) Xeon(R) CPU
E5-2620@2.00GHz, Ubuntu Linux 14.04, Kernel 3.13.0-30-
generic x86 64, GCC (Ubuntu 4.8.2-19ubuntu1), and
Python 2.7.6.

\subsection{Experiment one}
\label{subsec:Experiment-one}

In this experiment, we test the algorithms on polynomials. The grammar and the functions (taken from~\cite{Nguyen}) are shown, respectively, in Figure~\ref{fig:SCFG-1} and Table~\ref{tab:functions-first-experiment}.

\begin{figure}[htbp]
\begin{tcolorbox}[enhanced jigsaw, drop fuzzy shadow=ShadowColor, colback=gray!5]
{\footnotesize{}Exp~:=~Exp~+~Term $\mid$ Exp~-~Term $\mid$ Term $\mid$ }\textit{\footnotesize{}probs}{\footnotesize{}~0.33~0.33~0.33}{\footnotesize \par}
{\footnotesize{}Term~:=~Term~{*}~Factor~$\mid$~Term~/~Factor~$\mid$~Factor~$\mid$~}\textit{\footnotesize{}probs}{\footnotesize{}~0.33~0.33~0.33}{\footnotesize \par}
{\footnotesize{}Factor~:=~(Exp)~$\mid$~Num~$\mid$~X~$\mid$~}\textit{\footnotesize{}probs}{\footnotesize{}~0.33~0.33~0.33}{\footnotesize \par}
{\footnotesize{}Num~=:~1~$\mid$~2~$\mid$~3~$\mid$~4~$\mid$~5~$\mid$~6~$\mid$~7~$\mid$~8~$\mid$~9~$\mid$~}\textit{\footnotesize{}probs}{\footnotesize{}~0.11~0.11~0.11~0.11~0.11~0.11}
{\footnotesize{}~0.11~0.11~0.11}{\footnotesize \par}
{\footnotesize{}X~:=~$x_1 \mid ... \mid x_D\mid~$}\textit{\footnotesize{}probs}{\footnotesize{}~...}{\footnotesize \par}{\footnotesize \par}
\end{tcolorbox}
\caption{SCFG used in the first experiment. The \textit{X} rule depends on the dimension $D$ of the function (number of input data terminals).}
\label{fig:SCFG-1}
\end{figure}

\begin{table}[htbp]
\resizebox{0.8\columnwidth}{!}{
\begin{tabular}{c|c|c|c}
\hline 
\textbf{Name} & \textbf{Function} & \textbf{D} & \textbf{Train / Test}\tabularnewline
\hline 
\hline 
nguyen1 & $x^{3}+x^{2}+x$ & 1 & U{[}-1,1,20{]}\tabularnewline
nguyen2 & $x^{4}+x^{3}+x^{2}+x$ & 1 & U{[}-1,1,20{]}\tabularnewline
nguyen3 & $x^{5}+x^{4}+x^{3}+x^{2}+x$ & 1 & U{[}-1,1,20{]}\tabularnewline
nguyen4 & $x^{6}+x^{5}+x^{4}+x^{3}+x^{2}+x$ & 1 & U{[}-1,1,20{]}\tabularnewline
\hline 
\end{tabular}}
\caption{The SR functions used in the first experiment. 
	$U[a,b,c]$ means \textit{c} samples from a uniform distribution from \textit{a} to \textit{b}.}
\label{tab:functions-first-experiment}
\end{table}

\begin{table*}[!htbp]
\center%
\begin{tabular}{c|c|c|c|c|c|c|c|c}
\cline{1-9} 
\multirow{ 2}{*}{\textbf{\footnotesize{}Method}}
 & \multicolumn{2}{c|}{\textbf{\footnotesize{}\textbf{nguyen1}}} & \multicolumn{2}{c|}{\textbf{\footnotesize{}\textbf{nguyen2}}} & \multicolumn{2}{c|}{\textbf{\footnotesize{}\textbf{nguyen3}}} & \multicolumn{2}{c}{\textbf{\footnotesize{}\textbf{nguyen4}}}\tabularnewline
\cline{2-9} 
 & \textbf{\footnotesize{}MMAE (MAD)} & \textbf{\footnotesize{}Success} & \textbf{\footnotesize{}MMAE (MAD)} & \textbf{\footnotesize{}Success} & \textbf{\footnotesize{}MMAE (MAD)} & \textbf{\footnotesize{}Success} & \textbf{\footnotesize{}MMAE (MAD)} & \textbf{\footnotesize{}Success}\tabularnewline
\hline 
\hline 
\textbf{\textit{\footnotesize{}effmut}} & {\footnotesize{}0.20 (0.53)} & {\footnotesize{}0.32} & {\footnotesize{}0.22 (0.43)} & {\footnotesize{}0.23} & {\footnotesize{}0.27 (0.34)} & {\footnotesize{}0.11} & {\footnotesize{}0.24 (0.35)} & {\footnotesize{}0.05}\tabularnewline
\textbf{\footnotesize{}GB-LGP} & {\footnotesize{}2.77e-17 (0.34)} & {\footnotesize{}0.88} & {\footnotesize{}4.99e-17 (0.29)} & {\footnotesize{}0.8} & {\footnotesize{}0.07 (0.59)} & {\footnotesize{}0.33} & {\footnotesize{}0.08 (0.22)} & {\footnotesize{}0.04}\tabularnewline
\textbf{\footnotesize{}Hybrid v1} & {\footnotesize{}3.67e-17 (0.23)} & {\footnotesize{}0.86} & {\footnotesize{}6.00e-17 (4.58e-16 )} & {\footnotesize{}0.69} & {\footnotesize{}0.09 (0.39)} & {\footnotesize{}0.22} & {\footnotesize{}0.12 (0.29)} & {\footnotesize{}0.06}\tabularnewline
\textbf{\footnotesize{}Hybrid v2} & {\footnotesize{}3.26e-17 (2.24e-15)} & {\footnotesize{}0.74} & {\footnotesize{}0.02 (0.14)} & {\footnotesize{}0.5} & {\footnotesize{}0.09 (0.47)} & {\footnotesize{}0.27} & {\footnotesize{}0.14 (0.34)} & {\footnotesize{}0.05}\tabularnewline
\hline 
\end{tabular}
\caption{Descriptive statistics for experiment one.}
\label{tab:Median,-MAD,exp1}
\end{table*}

\begin{figure*}[ht]
	\center\includegraphics[scale=0.52]{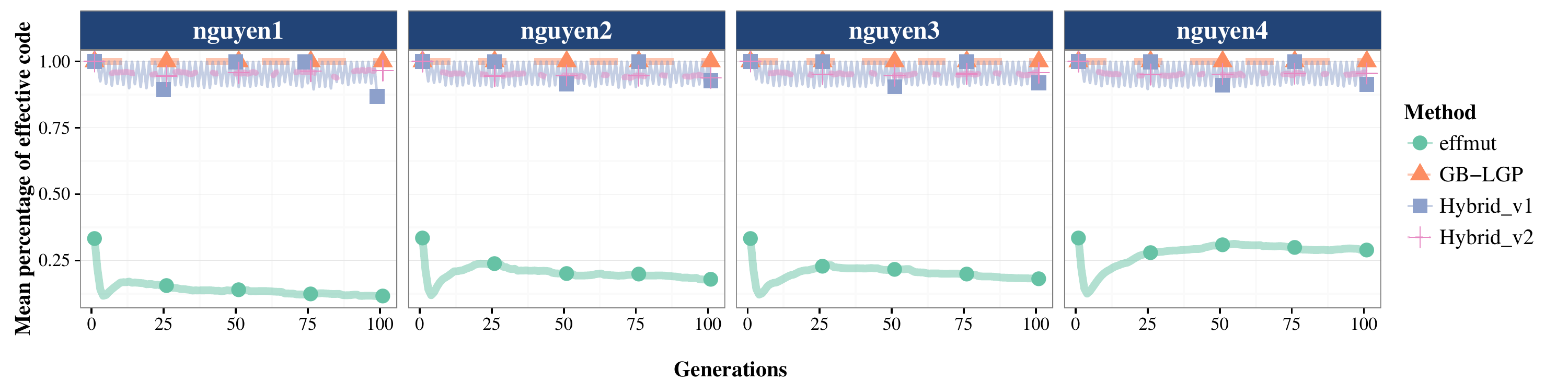}
	\caption{Mean percentage of effective code throughout the generations. The curves are means over 100 runs.}
	\label{fig:Mean-percentage-exp1}
\end{figure*}

Table \ref{tab:Median,-MAD,exp1} shows the results obtained for this experiment: MMAE, MAD, and success rate. Table \ref{tab:p-values-exp1} shows the \textit{p}-values from the statistical comparison.

\begin{table}[hbp]
\resizebox{\columnwidth}{!}{
\begin{tabular}{c|c|c|c|c}
\cline{1-5} 
{\textbf{\scriptsize{}Method}}
 & \textbf{\scriptsize{}nguyen1} & \textbf{\scriptsize{}nguyen2} & \textbf{\scriptsize{}nguyen3} & \textbf{\scriptsize{}nguyen4}\tabularnewline
\hline 
\hline 
\textbf{\scriptsize{}GB-LGP x effmut} & {\scriptsize{}5.17e-13 } & {\scriptsize{}2.72e-15 } & {\scriptsize{}6.63e-10 } & {\scriptsize{}2.69e-06 }\tabularnewline
\textbf{\scriptsize{}Hybrid v1 x effmut} & {\scriptsize{}8.88e-10 } & {\scriptsize{}1.18e-13 } & {\scriptsize{}1.56e-06 } & {\scriptsize{}6.00e-04}\tabularnewline
\textbf{\scriptsize{}Hybrid v2 x effmut} & {\scriptsize{}3.41e-10 } & {\scriptsize{}8.26e-09 } & {\scriptsize{}2.28e-06 } & {\scriptsize{}4.13e-03}\tabularnewline
\textbf{\scriptsize{}Hybrid v1 x GB-LGP} & {\scriptsize{}0.19} & {\scriptsize{}0.12} & {\scriptsize{}0.06} & {\scriptsize{}0.20}\tabularnewline
\textbf{\scriptsize{}Hybrid v2 x GB-LGP} & {\scriptsize{}0.19} & {\scriptsize{}3.15e-03} & {\scriptsize{}0.06} & {\scriptsize{}0.01}\tabularnewline
\textbf{\scriptsize{}Hybrid v1 x Hybrid v2} & {\scriptsize{}0.91} & {\scriptsize{}0.06} & {\scriptsize{}0.88} & {\scriptsize{}0.21}\tabularnewline
\hline 
\end{tabular}}
\caption{\textit{p}-values from Pairwise Wilcoxon Rank-Sum Test.}
\label{tab:p-values-exp1}
\end{table}

\begin{table*}[!htbp]
\center%
\begin{tabular}{c|c|c|c|c}
\hline 
\textbf{\footnotesize{}Name} & \textbf{\footnotesize{}Function} & \textbf{\footnotesize{}D} & \textbf{\footnotesize{}Train} & \textbf{\footnotesize{}Test}\tabularnewline
\hline 
\hline 
{\footnotesize{}\textbf{nguyen3}} & {\footnotesize{}$x^{5}+x^{4}+x^{3}+x^{2}+x$} & {\footnotesize{}1} & {\footnotesize{}U{[}-1,1,20{]}} & {\footnotesize{}U{[}-1,1,20{]}}\tabularnewline
{\footnotesize{}\textbf{nguyen6}} & {\footnotesize{}$sin(x)*sin(x+x^{2})$} & {\footnotesize{}1} & {\footnotesize{}U{[}-1,1,20{]}} & {\footnotesize{}U{[}-1,1,20{]}}\tabularnewline
{\footnotesize{}\textbf{keijzer4}} & {\footnotesize{}$x^{3}*e^{-x}*cos(x)*sin(x)*(sin(x)^{2}*cos(x)-1)$} & {\footnotesize{}1} & {\footnotesize{}E{[}0,10,0.05{]}} & {\footnotesize{}E{[}0.05,10.05,0.05{]}}\tabularnewline
{\footnotesize{}\textbf{keijzer5}} & {\footnotesize{}$\frac{30*x*z}{(x-10)*y^{2}}$} & {\footnotesize{}3} & {\footnotesize{}x,z: U{[}-1,1,500{]}} & {\footnotesize{}x,z: U{[}-1,1,10000{]}}\tabularnewline
 &  &  & {\footnotesize{}y: U{[}1,2,500{]}} & {\footnotesize{}y: U{[}1,2,10000{]}}\tabularnewline
{\footnotesize{}\textbf{korns3}} & {\footnotesize{}$-5.41+4.9*\frac{v-x+\nicefrac{y}{w}}{3*w}$} & {\footnotesize{}5} & {\footnotesize{}U{[}-50,50,500{]}} & {\footnotesize{}U{[}-50,50,10000{]}}\tabularnewline
{\footnotesize{}\textbf{korns5}} & {\footnotesize{}$3+2.13*ln(w)$} & {\footnotesize{}5} & {\footnotesize{}U{[}0,50,500{]}} & {\footnotesize{}U{[}0,50,10000{]}}\tabularnewline
\hline 
\end{tabular}
\caption{The SR functions used in the first experiment. $E[a,b,c]$ means a sequence of numbers spaced by an interval of \textit{c} from \textit{a} to \textit{b}.}
\label{tab:functions-second-experiment}
\end{table*}

\begin{table*}[!htbp]
\center%
\begin{tabular}{c|c|c|c|c|c|c}
\cline{1-7} 
{\textbf{\footnotesize{}Method}}
 & \multicolumn{1}{c|}{\textbf{\footnotesize{}nguyen3}} & \multicolumn{1}{c|}{\textbf{\footnotesize{}nguyen6}} & \multicolumn{1}{c|}{\textbf{\footnotesize{}keijzer4}} & \multicolumn{1}{c|}{\textbf{\footnotesize{}keijzer5}} & \multicolumn{1}{c|}{\textbf{\footnotesize{}korns3}} & \multicolumn{1}{c}{\textbf{\footnotesize{}korns5}}\tabularnewline
\hline 
\hline 
\textbf{\textit{\footnotesize{}effmut}} & \footnotesize{}0.223 (0.315) & \footnotesize{}0.083 (0.342) & \footnotesize{}0.204 (1.35e-16) & \footnotesize{}0.348 (0.089) & \footnotesize{}528.54 (0.27) & \footnotesize{}1.493 (0.077)\tabularnewline
\textbf{\footnotesize{}GB-LGP} & \footnotesize{}0.042 (0.184) & \footnotesize{}0.017 (0.245) & \footnotesize{}0.159 (0.028) & \footnotesize{}0.136 (0.063) & \footnotesize{}594.31 (0.09) & \footnotesize{}0.628 (0.363)\tabularnewline
\textbf{\footnotesize{}Hybrid v1} & \footnotesize{}0.048 (0.184) & \footnotesize{}0.024 (0.244) & \footnotesize{}0.154 (0.021) & \footnotesize{}0.134 (0.044) & \footnotesize{}291.29 (0.13) & \footnotesize{}0.778 (0.464)\tabularnewline
\textbf{\footnotesize{}Hybrid v2} & \footnotesize{}0.049 (0.207) & \footnotesize{}0.022 (0.178) &\footnotesize{} 0.182 (0.112) & \footnotesize{}0.147 (0.072) & \footnotesize{}666.26 (0.15) & \footnotesize{}1.275 (0.162)\tabularnewline
\hline 
\end{tabular}
\caption{Descriptive statistics - MMAE (MAD) - for experiment two.}
\label{tab:Median,-MAD,exp2}
\end{table*}

\begin{table*}[!htbp]
\center%
\begin{tabular}{c|c|c|c|c|c|c}
\cline{1-7} 
{\textbf{\footnotesize{}Method}}
 & \textbf{\footnotesize{}nguyen3} & \textbf{\footnotesize{}nguyen6} & \textbf{\footnotesize{}keijzer4} & \textbf{\footnotesize{}keijzer5} & \textbf{\footnotesize{}korns3} & \textbf{\footnotesize{}korns5}\tabularnewline
\hline 
\hline 
\textbf{\footnotesize{}GB-LGP x \textit{effmut}} & \footnotesize{}3.317e-21 & \footnotesize{}1.328e-14 & \footnotesize{}3.350e-15 & \footnotesize{}5.469e-12 & \footnotesize{}1 & \footnotesize{}6.612e-12\tabularnewline
\textbf{\footnotesize{}Hybrid v1 x \textit{effmut}} & \footnotesize{}1.644e-15 & \footnotesize{}1.229e-09 & \footnotesize{}1.141e-19 & \footnotesize{}1.569e-10 & \footnotesize{}0.774 & \footnotesize{}5.039e-14\tabularnewline
\textbf{\footnotesize{}Hybrid v2 x \textit{effmut}} & \footnotesize{}1.481e-16 & \footnotesize{}4.815e-08 & \footnotesize{}3.463e-12 & \footnotesize{}5.897e-06 & \footnotesize{}1 & \footnotesize{}1.086e-11\tabularnewline
\textbf{\footnotesize{}Hybrid v1 x GB-LGP} & \footnotesize{}0.025 & \footnotesize{}0.119 & \footnotesize{}3.326e-4 & \footnotesize{}0.960 & \footnotesize{}0.269 & \footnotesize{}0.894\tabularnewline
\textbf{\footnotesize{}Hybrid v2 x GB-LGP} & \footnotesize{}0.025 & \footnotesize{}0.097 & \footnotesize{}0.182 & \footnotesize{}0.134 & \footnotesize{}1 & \footnotesize{}0.832\tabularnewline
\textbf{\footnotesize{}Hybrid v1 x Hybrid v2} & \footnotesize{}0.884 & \footnotesize{}0.680 & \footnotesize{}7.257e-06  & \footnotesize{}0.134 & \footnotesize{}0.241 & \footnotesize{}0.832\tabularnewline
\hline 
\end{tabular}
\caption{\textit{p}-values from Pairwise Wilcoxon Rank-Sum Test.}
\label{tab:p-values-exp2}
\end{table*}

Compared to \textit{effmut}, the algorithms that employ the probabilistic model yielded much better results. For the simpler functions (\textit{nguyen1} and \textit{nguyen2}), the success rate of GB-LGP was at least 80\%, while \textit{effmut} reached a maximum of 32\%. For the fifth order polynomial (\textit{nguyen3}), the error increased substantially, but the proposed approaches still showed a much better success rate than \textit{effmut}. Finally, for the sixth order polynomial (\textit{nguyen4}), the success rate was very low and very similar for all techniques. However, the probabilistic algorithms found solutions with a much smaller error.

Given that the grammar is simple, and the rules that need to be learned are clear, the inferior performance on the higher order polynomials may be explained by the limitation of registers, which limit the depth of the equivalent tree. For instance, if the production \textit{Term := Factor} is chosen multiple times in sequence, although the result is not affected, the tree depth is, which limits the further representation of the program. A suggestion could be removing such sequences from the individuals, leaving only one of the equivalent productions. However, as we need them to know the productions used to generate the instructions, we must advance our research to elaborate a better tracking system.

As for the hybrid approaches, although they did perform better than \textit{effmut}, which was our primary objective, they did not outperform GB-LGP. The statistical tests in Table~\ref{tab:p-values-exp1} suggest that the differences between GB-LGP and the hybrids are not, in the most part, significant. However, one can observe a trend in Table~\ref{fig:Mean-percentage-exp1} for the hybrids being worse than GB-LGP. We do not know if the effects on the productions of the individuals caused by the mutation operators were the reason for this lower performance; it will be deeply investigated in a future work. Nevertheless, because the hybrids resample less than GB-LGP, they are faster and could be preferable over GB-LGP to solve these problems.

In order to asses the impact of the hybrid approach on the individuals, Figure~\ref{fig:Mean-percentage-exp1} shows the evolution of the mean percentage of effective code in the population for each function. As expected, individuals in \textit{effmut} have very few effective code. 
In GB-LGP, the entire code is effective, as no genetic operator is applied on them. Hybrid~v1 switches between having only effective code and having a little of non-effective code. As the population was resampled every two generations, there was no time to increase the amount of non-effective code. In Hybrid~v2, programs remained mostly effective, with only a small amount of non-effective code. We conclude that the hybrid approaches worked well in introducing LGP features into the programs.

\subsection{Experiment two}
\label{subsec:Experiment-two}

In this experiment, we incorporated some other functions to the grammar and tested the algorithms on different benchmark functions from \cite{Korns2011,Keijzer:ImprovingSR}. Figure \ref{fig:SCFG-2} shows the resulting grammar and Table \ref{tab:functions-second-experiment} specifies the functions.

\begin{table*}[!htbp]
	\center%
		\begin{tabular}{c|c|c|c|c|c|c}
			\cline{1-7} 
			{\textbf{\footnotesize{}Method}}
			& \multicolumn{1}{c|}{\textbf{\footnotesize{}nguyen3}} & \multicolumn{1}{c|}{\textbf{\footnotesize{}nguyen6}} & \multicolumn{1}{c|}{\textbf{\footnotesize{}keijzer4}} & \multicolumn{1}{c|}{\textbf{\footnotesize{}keijzer5}} & \multicolumn{1}{c|}{\textbf{\footnotesize{}korns3}} & \multicolumn{1}{c}{\textbf{\footnotesize{}korns5}}\tabularnewline
			\hline 
			\hline 			
			\textbf{\textit{\footnotesize{}effmut}} &\footnotesize{} 7.5 / 57 &\footnotesize{} 12 /58 &\footnotesize{} 9.5 / 48 &\footnotesize{} 13 / 51 &\footnotesize{} 13 / 51.5 &\footnotesize{} 5 / 57.5\tabularnewline
			\textbf{\footnotesize{}GB-LGP} &\footnotesize{} 21 / 21 &\footnotesize{} 22 / 22 &\footnotesize{} 35 / 35 &\footnotesize{} 29 / 29 &\footnotesize{} 29.5 / 29.5 &\footnotesize{} 21.5 / 21.5\tabularnewline
			\textbf{\footnotesize{}Hybrid v1} &\footnotesize{} 22 / 26 &\footnotesize{} 25 / 28.5 &\footnotesize{} 34 / 44.5 &\footnotesize{} 29.5 / 40.5 &\footnotesize{} 35.5 / 43.5 &\footnotesize{} 20.5 / 24\tabularnewline
			\textbf{\footnotesize{}Hybrid v2} &\footnotesize{} 25 / 25 &\footnotesize{} 23 / 25 &\footnotesize{} 28 / 29 &\footnotesize{} 31.5 / 36.5 &\footnotesize{} 33 / 34.5 &\footnotesize{} 21.5 / 24.5\tabularnewline
			\hline 
		\end{tabular}
		\caption{Mean effective size / total size of the final solution.}
		\label{tab:effective-sizes-exp2}
	\end{table*}

MMAE and MAD for this second experiment are shown in Table~\ref{tab:Median,-MAD,exp2}, and the \textit{p}-values are shown in Table~\ref{tab:p-values-exp2}. We are not presenting the success rates because they were very small. 

The results in this experiment follow the same pattern of the previous experiment, with GB-LGP yielding the best results, and the techniques that use the probabilistic model performing better than \textit{effmut}. As both the functions and the grammar are more complex in this experiment, the prediction errors (MMAE and MAD) were not as small as in the previous experiment. No method could perform well on \textit{Korns'} functions, likely due to the constants that must be tuned.

\begin{figure}[htpb]
	
%

\begin{tcolorbox}[enhanced jigsaw, drop fuzzy shadow=ShadowColor, colback=gray!5, left=8pt]
	{\footnotesize{}Exp~:=~Exp~+~Term $\mid$ Exp~-~Term $\mid$ Term $\mid$ }\textit{\footnotesize{}probs}{\footnotesize{}~0.33~0.33~0.33}{\footnotesize \par}
	{\footnotesize{}Term~:=~Term~{*}~Factor~$\mid$~Term~/~Factor~$\mid$~Factor~$\mid$~}\textit{\footnotesize{}probs}{\footnotesize{}~0.33~0.33~0.33}{\footnotesize \par}
	{\footnotesize{}Factor~:=~sin(Arg)~$\mid$~cos(Arg)~$\mid$~exp(Arg)~$\mid$~ln(Arg)~$\mid$~Arg~}$\mid$ \textit{\footnotesize{}probs}{\footnotesize{}~0.2~0.2~0.2~0.2~0.2}{\footnotesize \par}
	{\footnotesize{}Arg~:=~(Exp)~$\mid$~Num~$\mid$~X~$\mid$}\textit{\footnotesize{}probs}{\footnotesize{}~0.33~0.33~0.33}{\footnotesize \par}{\footnotesize \par}
	{\footnotesize{}Num~=:~1~$\mid$~2~$\mid$~3~$\mid$~4~$\mid$~5~$\mid$~6~$\mid$~7~$\mid$~8~$\mid$~9~$\mid$~}\textit{\footnotesize{}probs}{\footnotesize{}~0.11~0.11~0.11~0.11~0.11~0.11}
	{\footnotesize{}~0.11~0.11~0.11}{\footnotesize \par}
	{\footnotesize{}X~:=~$x_1 \mid ... \mid x_D\mid~$}\textit{\footnotesize{}probs}{\footnotesize{}~...}{\footnotesize \par}{\footnotesize \par}
\end{tcolorbox}

\caption{SCFG used in the second experiment. The \textit{X} rule depends on the dimension $D$ of the benchmark function (number of terminals).}
\label{fig:SCFG-2}
\end{figure}

Although the mean error shown in Table \ref{tab:Median,-MAD,exp2} for GB-LGP on \textit{nguyen3} is lower than that in Table \ref{tab:Median,-MAD,exp1}, in this second experiment the algorithm had no success. The population approximated the data with other functions present in the extended grammar (Figure~\ref{fig:SCFG-2}), such as \textit{sin} and \textit{cos}. This suggests that a more complex grammar may lead the search more easily to stagnation on sub-optimal (wrong) solutions.

The hybrid versions were expected to overcome that issue at some level, but it did not occur. The failure may be related to the parameters or the way the productions associated with an individual are updated after a mutation. Nonetheless, the use of the hybrid versions make it possible to obtain better results than with \textit{effmut} in less time than with GB-LGP, as less samplings from the grammar take place.

\begin{figure*}[htbp]
\center\includegraphics[width=0.85\textwidth]{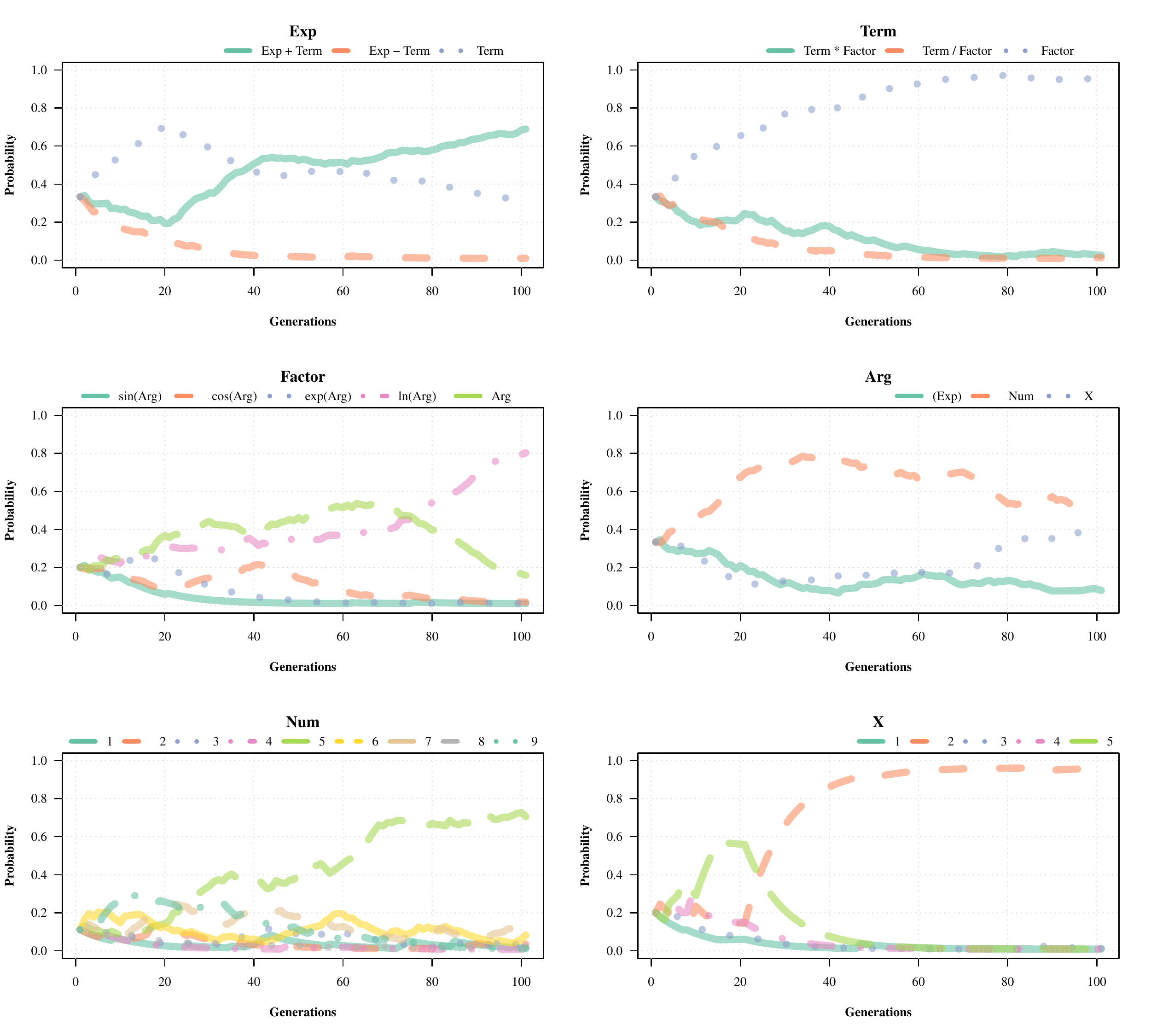}
\caption{Evolution of the probability distribution of each rule of the grammar throughout the generations for \textit{Korns5}. The curves are the mean of the 100 independent runs.}
\label{fig:Evolution-korns5}
\end{figure*}

In order to evaluate how big solutions get using the grammar, Table \ref{tab:effective-sizes-exp2} shows the mean size, in number of instructions and effective instructions, of the solutions found. The effective size of the solutions created with the grammar are not much bigger than the ones found by \textit{effmut}. In fact, they have a smaller total size.

It is not possible to show how every grammar evolved for each technique on each function, due to the large amount of space needed for that. However, to illustrate how the grammar evolves, Figure~\ref{fig:Evolution-korns5} shows the evolution of the probability distributions for GB-LGP on the \textit{Korns5} function. One can observe, for instance, that the high probability of generating solutions containing \textit{ln}, a number (5), and $x_{2}$, which is the only input that the benchmark function uses. This plot is a strong indicative that our algorithms are working as expected and that  improvements can be obtained in further investigations. 

\section{Conclusions and Future Work}
\label{sec:Conclusion-and-Future}

The use of grammars to enhance the performance of Genetic Programming algorithms is a well-investigated topic, but its Linear counterpart was still unexplored, as far as we know. LGP is harder to implement than GP, and tracking the changes to adequately update the model proved to not be trivial. 

In this work, the use of SCFG to acquire some knowledge about the search space during the evolution and to guide the process allowed the proposed algorithms to obtain better results than the standard LGP on a set of well-known Symbolic Regression problems. When using a simple grammar on polynomials, GB-LGP was able to achieve high success rates. The hybrid approaches were also able to outperform the standard LGP (\textit{effmut}) while introducing LGP mutations and reducing the execution time. The promising results open an avenue for many future investigations.

The proposed algorithms faced some issues: the existence of sequences of instructions that represent the same production rule and do not affect the final result of a program, like an identity attribution $r[0]=r[0]$ used several times in sequence; the effects of the mutation operators to the productions associated to an individual; the convergence to local optima, or simply a stagnation in the wrong regions of the search-space, because of function approximations using the wrong operators.

In future works, the following issues will be explored:
\begin{itemize}
\item Removing sequences of equivalent instructions that augment the depth of the equivalent tree;
\item Investigate parsing the individuals after the use of a mutation operator in order to correctly find which productions from the grammar should
be updated;
\item Develop genetic operators that take into consideration information of the probability distributions of the SCFG rules;
\item Investigate methods to reduce the complexity (size) of the solutions;
\item Either incorporate dependency into SCFG or replace it by another model better fit to the LGP representation.
\end{itemize}



\begin{acks}

This work was supported by Funda\c{c}\~{a}o de Amparo \`{a} Pesquisa do Estado
de S\~{a}o Paulo (FAPESP \textendash{} Proc. 2016/07095-5).

\end{acks}